%% file: main.tex
\documentclass[10pt,twocolumn,letterpaper]{article}

\usepackage{cvpr}              
\input{preamble}
\definecolor{cvprblue}{rgb}{0.21,0.49,0.74}
\usepackage[pagebackref,breaklinks,colorlinks,allcolors=cvprblue]{hyperref}
\usepackage{multirow,eucal}
\def\paperID{5700} 
\def\confName{CVPR}
\def\confYear{2026}

\title{Exploring Spatial Intelligence from a Generative Perspective}

\author{
\begin{tabular}{cccc}
    Muzhi Zhu$^{1,2*}$ ~~
    Shunyao Jiang$^{1*}$ ~~
    Huanyi Zheng$^{1}$ ~~
    Zekai Luo$^{1}$ ~~
    Hao Zhong$^{1}$ ~~
    Anzhou Li$^{1,2}$ \\
    Kaijun Wang$^{1}$ ~~
    Jintao Rong$^{4}$ ~~
    Yang Liu$^{1}$ ~~
    Hao Chen$^{1\dagger}$ ~~
    Tao Lin$^{3,2}$ ~~
    Chunhua Shen$^{1,2\dagger}$
\end{tabular} \\[.25cm]
\small{$^1$Zhejiang University, State Key Laboratory of CAD \& CG \quad $^2$Ant Group  \quad $^3$Westlake University \quad $^4$Zhejiang University of Technology}
}
\newcommand\blfootnote[1]{%
    \begingroup
    \renewcommand\thefootnote{}\footnote{#1}%
    \addtocounter{footnote}{-1}%
    \endgroup
}
\begin{document}
\maketitle
\blfootnote{$^*$ Equal contribution. $^\dagger$ Corresponding authors.}
\input{sec/0_abstract}    
\input{sec/1_intro}
\input{sec/2_related_work}
\input{sec/3_method}

\input{sec/4_experiments}

\input{sec/5_conclusion}%

\section*{Acknowledgments}
This work was supported in part by The Pioneer R\&D Program of Zhejiang (Grant No.~2025C01011), by the Ant Group Research Intern Program, and by the National Natural Science Foundation of China (Grant No.~62576315).

{
    \small
    \bibliographystyle{ieeenat_fullname}
    \bibliography{main}
}

\end{document}

%% file: sec/0_abstract.tex
\begin{abstract}

Spatial intelligence is essential for multimodal large language models, yet current benchmarks largely assess it only from an \emph{understanding} perspective. We ask whether modern generative or unified multimodal models also possess \emph{generative spatial intelligence} (GSI)—the ability to respect and manipulate 3D spatial constraints during image generation—and whether such capability can be \emph{measured} or \emph{improved}. We introduce \textbf{GSI-Bench}, the first benchmark designed to quantify GSI through spatially grounded image editing. 
It consists of two complementary components: \textbf{GSI-Real}, a high-quality real-world dataset built via a 3D-prior-guided generation and filtering pipeline, and \textbf{GSI-Syn}, a large-scale synthetic benchmark with controllable spatial operations and fully automated labeling.
Together with a unified evaluation protocol, GSI-Bench enables scalable, model-agnostic assessment of spatial compliance and editing fidelity. Experiments show that fine-tuning unified multimodal models on GSI-Syn yields substantial gains on both synthetic and real tasks and, strikingly, also improves downstream spatial \emph{understanding}. This provides the first clear evidence that generative training can tangibly strengthen spatial reasoning—establishing a new pathway for advancing spatial intelligence in multimodal models.
\begin{figure*}[htbp]
    \centering
    \includegraphics[width=1\linewidth]{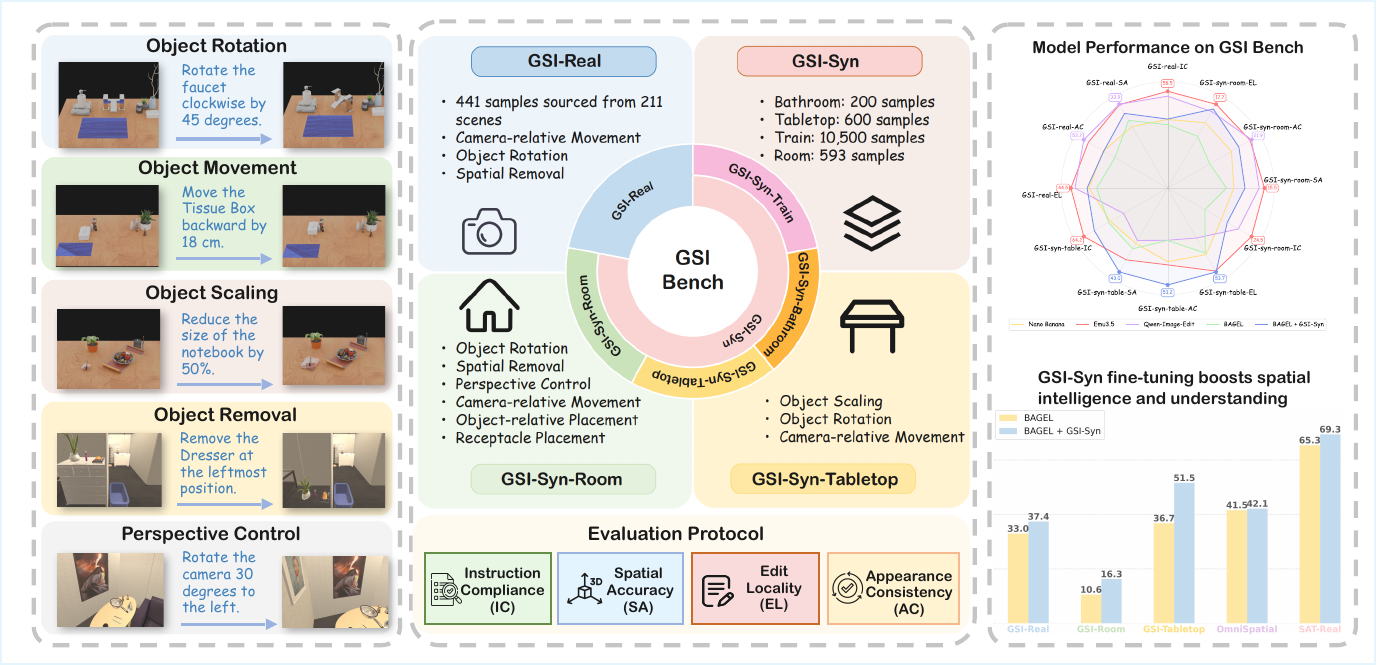}
    \caption{We introduce GSI Bench, a benchmark for grounded spatial intelligence that spans both real-world and synthetic scenes. GSI Bench evaluates a diverse set of spatial editing skills across multiple domains. By incorporating fine-grained evaluation protocols covering instruction compliance, spatial accuracy, edit locality, and appearance consistency, GSI Bench enables rigorous assessment of spatial reasoning in image-editing models. We further show that fine-tuning with GSI-Syn significantly boosts models’ spatial understanding and generalization across all subsets of the benchmark.}
\label{fig:main}
\vspace{-4mm}
\end{figure*}
\end{abstract}

%% file: sec/1_intro.tex
\section{Introduction}
\label{sec:intro}

Spatial intelligence~\cite{jia2025omnispatial,cai2025has,yang2025thinking,yin2025mindcube,chen2024spacemantis}—the capacity to reason about objects, scenes, and their geometric relationships in the real 3D physical world—is foundational for multimodal large language models (MLLMs)~\cite{achiam2023gpt4,openai2024gpt4o,wang2024qwen2vl,chen2024spacemantis,wang2025genspace,blip3-o,comanici2025gemini,yang2025qwen3,Spatial-mllm}. It governs how models ground language in space and interact with the physical world, and is indispensable for embodied navigation~\cite{embodiednav,RoboTrom-Nav,embodiednfm}, robotic manipulation~\cite{huang2025notvla, ji2025robobrain, yuan2024robopoint}, and 3D scene understanding~\cite{yin2025mindcube, cui2025emu3,dai2017scannet,reizenstein2021common} under partial observability and domain shift. 
Despite this centrality, the prevailing ecosystem of datasets~\cite{Scannet++,jia2025omnispatial}, benchmarks~\cite{yang2025thinking,yin2025mindcube}, and modeling choices~\cite{achiam2023gpt4, wang2024emu3} has developed spatial intelligence predominantly from an \emph{understanding} perspective: recognition- or QA-style supervision, 2D/3D perception pipelines, and offline diagnostics on curated test suites. 

Meanwhile, a parallel trend has emerged: unified multimodal models~\cite{chen2025janus,ma2025janusflow,xie2024show,xie2025show,cui2025emu3,wang2024emu3,lin2025uniworld,wu2025omnigen2,blip3-o,deng2025emerging} that \emph{jointly} perform understanding and generation, aiming to demonstrate the mutual benefits between the two. Existing evidence largely confirms that stronger visual understanding can enhance image generation quality~\cite{wang2024qwen2vl,wang2025genspace,blip3-o,jia2025omnispatial}. Yet the reverse direction remains underexplored—can \emph{generation} itself help models acquire a deeper grasp of spatial concepts and thereby strengthen their \emph{understanding}? We argue that spatial intelligence offers a principled lens through which to investigate this question.

This paper takes a generative perspective on spatial intelligence. We ask: (1) Do modern generative or unified multimodal models exhibit \emph{generative spatial intelligence} (GSI)—the capacity to respect and manipulate spatial constraints during image generation? (2) Can GSI be \emph{measured} in a reliable, scalable, and model-agnostic way? (3) Can we \emph{enhance} GSI via targeted interventions, and does such enhancement transfer to downstream spatial \emph{understanding} tasks?

Such \emph{generative spatial intelligence} is not only crucial for generating and editing images~\cite{jiang2025anyedit,zhao2024ultraedit,gemini2023anil,comanici2025gemini} that faithfully preserve real-world spatial relationships, but also serves as a bridge connecting unified understanding–generation models with emerging paradigms such as ``thinking with images''~\cite{yang2025thinking, yin2025mindcube,Thinkingwithimagessurvey} and world models~\cite{ha2018world,LWM,Realworldsim,lin2025uniworld}. This connection provides a foundational step toward deploying these models in embodied and interactive real-world tasks such as navigation and manipulation.
To this end, we operationalize \emph{generative spatial intelligence} through a spatially grounded image editing task, where a unified multimodal model receives an input image and an unambiguous, spatially-related editing instruction, and is required to generate an output image that satisfies the specified spatial constraints. Constructing datasets and automated evaluation pipelines that accurately reflect such precise spatial concepts is highly non-trivial. We address this challenge from both real-world and synthetic perspectives.

For real scenes, the key advantage lies in the small domain gap to downstream applications, making them naturally aligned with embodied and perception-based tasks. However, they also pose inherent challenges: existing datasets~\cite{Scannet++,jia2025omnispatial,ray2024sat} rarely contain precise annotations for spatial manipulations, and it is often difficult to express the spatial operations between image pairs using clear, unambiguous natural language descriptions that humans can easily understand. To overcome this, we design a complete data generation and filtering pipeline that leverages 3D grounding priors and rule-based spatial operation generation, uses MLLMs as captioners and validators, and incorporates human verification. This process results in \textbf{GSI-Real}, the first high-quality real-world benchmark for spatially grounded image editing. Nevertheless, real-world data remains limited in scale and diversity. To complement it, we construct a large-scale synthetic benchmark, \textbf{GSI-Syn}, based on simulation environments~\cite{Ai2-thor,mesatask,InternScenes} with controllable rendering. GSI-Syn provides abundant, precisely labeled image pairs with diverse types and difficulty levels of spatial operations, and also offers an automated data generation pipeline for potential training use.


Finally, we fine-tune existing unified multimodal models on GSI-Syn and evaluate them on both synthetic and real benchmarks. Our experiments demonstrate that \emph{generative spatial intelligence} can be effectively enhanced through simulation-based training, and such improvements consistently transfer to spatial \emph{understanding} tasks.

In summary, our main contributions are as follows:
1) We introduce \textbf{GSI-Bench}, a comprehensive benchmark that operationalizes \emph{generative spatial intelligence} through spatially grounded image editing, enabling unified models to reason about and manipulate spatial relations during generation. 2) We construct two complementary components of \textbf{GSI-Bench}: \textbf{GSI-Real}, the first high-quality real-world dataset for spatially grounded editing, and \textbf{GSI-Syn}, a large-scale synthetic dataset and benchmark with controllable spatial operations and difficulty levels. 3) We establish an automated pipeline for dataset generation and evaluation that leverages 3D grounding priors, rule-based operation generation, multimodal captioning, and human verification. 4) We empirically demonstrate that simulation-based fine-tuning on GSI-Syn enhances \emph{generative spatial intelligence} and further improves downstream spatial understanding tasks. 



%% file: sec/2_related_work.tex
\section{Related Work}
\label{sec:related_work}

\subsection{Spatial Intelligence in MLLMs}
Spatial intelligence serves as a critical bridge connecting multimodal large language models to the physical 3D world.
However, existing research has primarily focused on the \emph{understanding} aspect of spatial reasoning, with limited exploration of \emph{generative} spatial capabilities.
On the benchmark side, recent efforts have probed MLLMs' spatial understanding from various perspectives. 
VSI-Bench~\cite{yang2025thinking} evaluates video-based spatial reasoning over temporal sequences.
MindCube~\cite{yin2025mindcube} examines 3D spatial modeling from sparse multi-view observations.
OmniSpatial~\cite{jia2025omnispatial} provides a systematic assessment across multiple spatial reasoning dimensions, including dynamic reasoning, spatial interaction, and perspective taking.
On the methodology side, several works~\cite{yuan2024robopoint,ji2025robobrain,yin2025mindcube} aim to enhance spatial understanding in MLLMs. 
Spatial-MLLM~\cite{Spatial-mllm} introduces an auxiliary spatial encoder to explicitly inject 3D geometric information into the model.
SAT~\cite{ray2024sat} leverages simulation environments to generate large-scale rule-based spatial reasoning data for training (real-world evaluation: SAT-Real).
REVISION~\cite{chatterjee2024revision} demonstrates that data from simulated rendering engines (e.g., Blender) can benefit both image generation and spatial understanding when used as additional guidance.
Despite these advances, prior work has not explored spatial intelligence from a unified understanding-generation perspective. This work pioneers the evaluation of generative spatial intelligence in unified MLLMs, showing that fine-tuning on spatial editing tasks improves spatial reasoning in both modalities.

\subsection{Unified Multimodal Models}

Recently, unified multimodal models for both image understanding and image generation have made rapid progress. Among {closed-source} systems, GPT-Image~\cite{openai2024gpt4o} integrates image generation directly into autoregressive language modeling, enabling attribute binding, text rendering, and iterative controlled editing within a unified token space. NanoBanana~\cite{comanici2025gemini} further emphasizes spatially controllable generation, supporting multi-image conditioning, localized editing, and pose/object manipulation while preserving structural and geometric consistency.
Meanwhile, the {open-source} community~\cite{chen2025janus,ma2025janusflow,xie2024show,xie2025show,blip3-o,deng2025emerging} is actively advancing the paradigm of a single model that unifies understanding and generation.
BAGEL~\cite{deng2025emerging} employs a Mixture-of-Transformers structure and achieves competitive performance on both vision understanding and generation. 
Emu3~\cite{wang2024emu3} introduces native multimodal next-token prediction, and Emu3.5~\cite{cui2025emu3} further extends this to interleaved image–text input/output, demonstrating capabilities in long-horizon scene modeling.
Despite these advances, existing unified models still lack systematic evaluation of spatial understanding and controllable editing  capabilities. 
To address this gap, we systematically benchmark multimodal models for their Generative Spatial Intelligence capability, providing the first comprehensive evaluation framework that connects generative and understanding aspects of spatial reasoning.


%% file: sec/3_method.tex





\section{Generative Spatial Intelligence}
\label{sec:definition}

\subsection{What is Generative Spatial Intelligence?}
\label{subsec:gsi-definition}

We define \emph{Generative Spatial Intelligence (GSI)} as the capability of a unified multimodal model to \textbf{respect, reason about, and manipulate spatial constraints during image generation}. In contrast to traditional spatial understanding---which focuses on perceiving or describing spatial configurations---GSI reflects whether a model can \emph{actively enforce} spatial relationships when generating new visual content.

Ideally, text-to-image generation could also manifest certain aspects of GSI, since generating a scene from a spatially descriptive prompt inherently requires reasoning about object layouts and relations. However, such setups typically lack sufficient constraints for precise assessment: the open-ended nature of text prompts introduces ambiguity, and there is no unique ground-truth target against which spatial consistency can be objectively measured. 
To more faithfully and quantitatively capture GSI, we therefore adopt an \emph{image-to-image editing} formulation. In this setting, the model receives both a reference image and a spatially grounded instruction, and must produce an edited image that satisfies the specified spatial transformation. This task demands not only understanding the spatial structure of the input image but also manipulating it coherently according to the instruction—thus directly revealing the model’s generative spatial reasoning capability.


\subsection{Task Formulation}
\label{subsec:task}

We operationalize GSI through a \emph{spatially grounded image editing} task that emphasizes \textbf{quantitative, controllable, and physically grounded spatial transformations}. 
Formally, given an input image $\mathcal{I}$ and a textual instruction $\mathcal{T}$ specifying a spatial manipulation, the model is required to generate an output image $\mathcal{I}' = f(\mathcal{I}, \mathcal{T})$ that accurately satisfies the intended transformation while maintaining realism and semantic consistency.

Different from prior qualitative editing tasks that focus on semantic or stylistic changes (e.g., ``make it look sunny''), our formulation introduces a suite of \textbf{quantitative spatial operations} that explicitly modify the underlying scene geometry rather than only pixel appearance. 
To formalize these operations, we first model each visual scene through its latent 3D structure, which defines object layouts, camera parameters, and their geometric relationships. 
This abstraction enables us to describe spatial manipulations as structured 3D transformations that can be consistently reflected in the generated image.

\noindent\textbf{3D Scene Representation.}
We represent each scene as $\mathcal{S} = \{\mathcal{O}_i\}_{i=1}^{N} \cup \{\mathcal{C}\}$, where $\mathcal{O}_i = (\mathbf{c}_i, \mathbf{s}_i, \mathbf{R}_i)$ denotes the $i$-th object with center $\mathbf{c}_i \in \mathbb{R}^3$, size $\mathbf{s}_i \in \mathbb{R}^3$, and rotation $\mathbf{R}_i \in SO(3)$, and $\mathcal{C} = (\mathbf{R}_c, \mathbf{t}_c, K)$ denotes the camera.
Any 3D point $\mathbf{p}_i$ can be projected to the image plane as $\tilde{\mathbf{p}}_i = \pi(K(\mathbf{R}_c \mathbf{p}_i + \mathbf{t}_c))$, establishing the geometric foundation for spatial manipulations and evaluation.

\noindent\textbf{Spatial Operation Representation.}
Each spatial instruction is structured as $\mathcal{T} = \langle \mathcal{R}, \mathcal{A}, \Phi_{\text{3D}} \rangle$, where $\mathcal{R}$ identifies target objects, $\mathcal{A}$ specifies the action, and $\Phi_{\text{3D}}: \mathcal{S}_{\text{src}} \rightarrow \mathcal{S}_{\text{dst}}$ defines the geometric transformation by updating object poses or camera parameters:
$(\mathbf{c}_i, \mathbf{R}_i, \mathbf{R}_c, \mathbf{t}_c)_{\text{src}} \mapsto (\mathbf{c}_i', \mathbf{R}_i', \mathbf{R}_c', \mathbf{t}_c')_{\text{dst}}$.
For instance, "Move the apple 15 cm left" induces a camera-relative translation, while "Place the cup left of the plate" defines a relational constraint $\mathbf{c}_{\text{cup}}' = \mathbf{c}_{\text{plate}} + \Delta_{\text{left}}$.
This formulation explicitly links linguistic spatial instructions with 3D geometric transformations, providing a unified interface for data synthesis, model training, and quantitative evaluation in GSI-Bench.
\vspace{-1mm}

\subsection{Categories of Spatial Operations}
\label{subsec:ops}
\vspace{-1mm}
We define seven quantitatively grounded spatial operations spanning object-, camera-, and scene-level transformations, enabling comprehensive GSI evaluation. 
Formal mathematical definitions are provided in the appendix material.

\begin{table}[t]
\vspace{-1mm}
\centering
\scriptsize
\setlength{\tabcolsep}{3.5pt}
\renewcommand{\arraystretch}{1.1}
\caption{Spatial operation taxonomy.}
\vspace{-1mm}
\label{tab:operations}
\begin{tabular}{@{}llp{4.8cm}@{}}
\toprule
\textbf{ID} & \textbf{Operation} & \textbf{Description \& Capability Tested} \\
\midrule
CM & Camera-Relative Move & Move along camera axes; egocentric reasoning \\
OP & Object-Relative Place & Position relative to reference; pairwise relations \\
OR & Object Rotation & Rotate objects; 3D orientation control \\
RP & Receptacle Placement & Place in containers; hierarchical reasoning \\
PC & Perspective Control & Change viewpoint; view-aware adaptation \\
SR & Spatial Removal & Delete by criteria; global spatial understanding \\
OS & Object Scaling & Scale uniformly; metric reasoning \\
\bottomrule
\end{tabular}
\vspace{-3mm}
\end{table}

\begin{figure*}[htbp]
    \centering
    \includegraphics[width=1\linewidth]{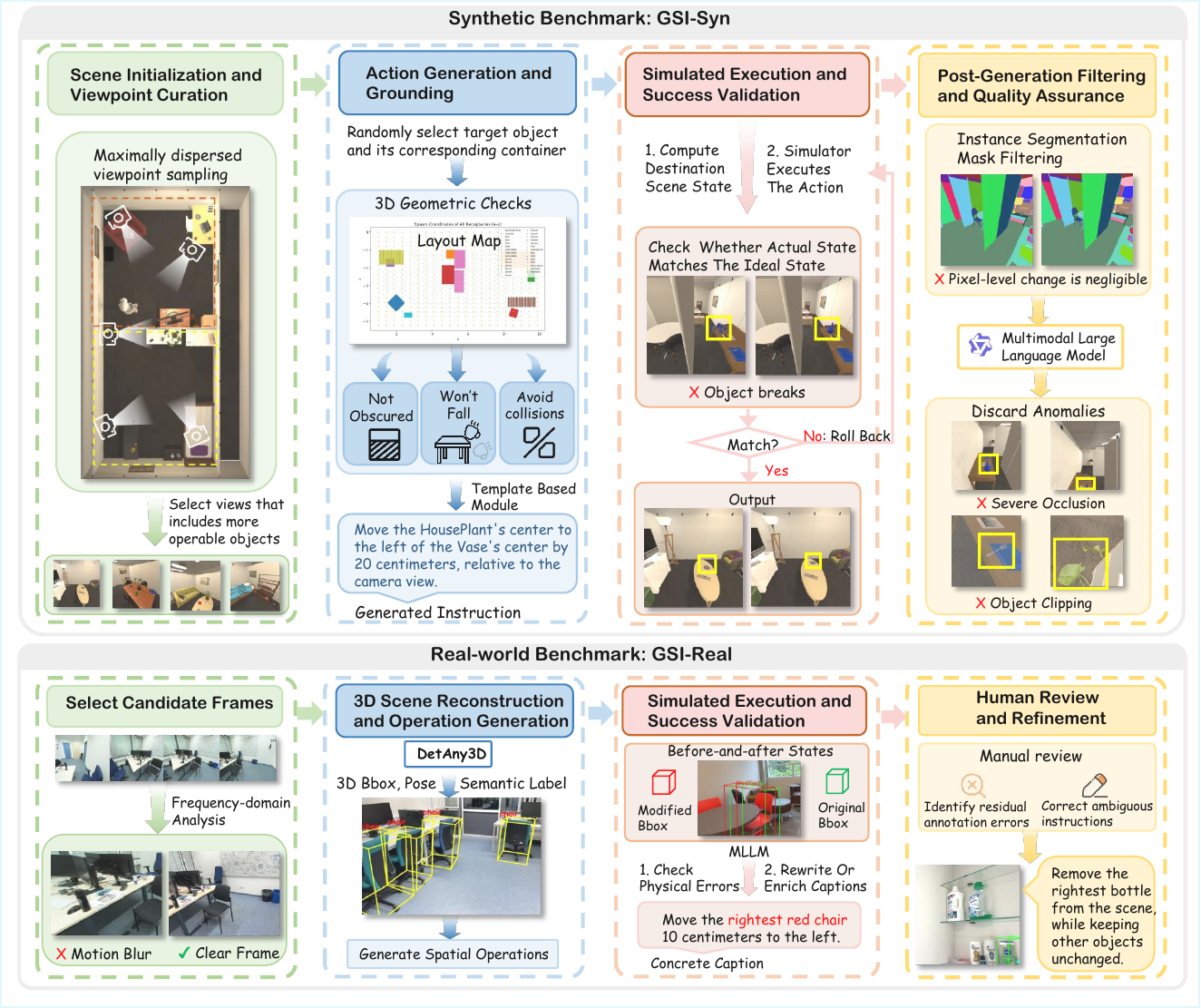}
    \caption{\textbf{Benchmark curation pipeline.}The pipeline builds both synthetic (GSI-Syn) and real-world (GSI-Real) benchmarks through unified scene processing, action generation, and validation. For GSI-Syn, scenes are sampled from diverse viewpoints, feasible actions are generated via 3D geometric checks, and a simulator validates outcomes before filtering failures and anomalies. For GSI-Real, clear frames are selected, 3D scene structure is reconstructed, and spatial operations are generated and validated on bounding boxes. Human review then refines captions and corrects residual annotation errors, ensuring high-quality spatial-editing supervision.}
\vspace{-2mm}
\end{figure*}

\section{GSI-Bench Construction}
\label{sec:benchmark}

\subsection{Synthetic Benchmark: GSI-Syn}
\label{subsec:syn-benchmark}

To facilitate scalable and controllable evaluation, we construct \textbf{GSI-Syn}, a large-scale synthetic benchmark for generative spatial intelligence. GSI-Syn is built upon open-source simulators including AI2-THOR~\cite{Ai2-thor} and MesaTask~\cite{mesatask}, covering varied scenarios like indoor navigation and tabletop manipulation. 
The primary advantage of this simulation-based approach is two-fold. First, it provides perfect ground-truth data, including the initial 3D scene representation ($\mathcal{S}_{\text{src}}$), the precise geometric transformation ($\Phi_{\text{3D}}$), and the resulting target scene ($\mathcal{S}_{\text{dst}}$), allowing for unambiguous, automated validation. Second, we can render the ground-truth edited image ($\mathcal{I}'$) directly from $\mathcal{S}_{\text{dst}}$, yielding high-quality $(\mathcal{I}, \mathcal{T}, \mathcal{I}')$ triplets for both evaluation and training. Our automated synthesis pipeline consists of the following stages.

\noindent\textbf{Scene Initialization and Viewpoint Curation.}
A key aspect of our data generation is sampling diverse and meaningful camera viewpoints. For each indoor scene, we employ DBSCAN clustering~\cite{DBSCAN} on the floor plan to partition the space into distinct rooms. Within each room, we perform maximally dispersed viewpoint sampling. To ensure these viewpoints are "actionable," we prioritize those containing more manipulable objects, guaranteeing each viewpoint can support a rich set of potential spatial operations.

\noindent\textbf{Action Candidate Generation and Geometric Grounding.}
For each viewpoint, we generate valid action candidates through object selection and multi-level geometric validation. We randomly select a target object, ensuring it is not occluded and rests on a stable surface. For relational operations (e.g., "place the apple to the left of the bowl"), a reference or container object is also selected. We then perform rigorous 3D geometric checks to verify physical plausibility: camera-relative translations are validated by ensuring the target remains visible and does not fall off its supporting surface; object-relative placements are checked for spatial sufficiency and collision avoidance. A template-based module generates the corresponding textual instruction $\mathcal{T}$.

\noindent\textbf{Simulated Execution and Success Validation.}
With a valid instruction $\mathcal{T}$ and transformation $\Phi_{\text{3D}}$, we execute the action in the physics-enabled simulator. We first analytically compute the \emph{ideal} destination state $\mathcal{S}_{\text{dst}}^{\text{ideal}}$, then the physics engine executes the action to produce the \emph{actual} outcome $\mathcal{S}_{\text{dst}}^{\text{actual}}$. An operation succeeds only if the actual state matches the ideal state, confirmed by checking the final position and visibility of the target object. Failed executions (e.g., due to unforeseen collisions) are rolled back and resampled.

\noindent\textbf{Post-Generation Filtering and Quality Assurance.}
To ensure benchmark quality, we apply two-stage filtering. First, using instance segmentation masks, we filter out samples where pixel-level change is negligible, ensuring every edit is visually significant. Second, we leverage an MLLM (Qwen3-VL-235B) as a quality gate to identify and discard samples with subtle anomalies difficult to capture with hard-coded rules, such as simulation artifacts (e.g., object clipping), physically implausible outcomes, or severe occlusions that render instructions ambiguous.

\medskip
Through this automated pipeline, GSI-Syn generates diverse, physically valid, and geometrically precise editing pairs at scale, offering a reproducible and extensible platform for probing spatial reasoning in generative models under fully controlled conditions.

\subsection{Real-world Benchmark: GSI-Real}
\label{subsec:real-benchmark}

To complement the synthetic GSI-Syn, we curate \textbf{GSI-Real}, a real-world benchmark for evaluating generative spatial intelligence in natural images.
Unlike simulation-based GSI-Syn, constructing GSI-Real presents unique challenges: we cannot obtain perfect 3D scene representations nor directly execute physical transformations to acquire ground-truth edited images ($\mathcal{I}'$). Consequently, we develop an alternative evaluation protocol that bypasses the need for $\mathcal{I}'$. Each sample in GSI-Real is represented as $(\mathcal{I}, \mathcal{T}, \mathcal{S}_{\text{src}}, \Phi_{\text{3D}}, \mathcal{S}_{\text{dst}})$, where the edited image is generated by the model under test, and success is evaluated by analyzing spatial consistency between the predicted edit and the specified 3D transformation.

\noindent\textbf{Image Source and Frame Selection.}
We source real-world images from ScanNet++~\cite{Scannet++}, a large-scale indoor RGB-D dataset. To ensure diversity and visual quality, we sample one frame from every 20 frames and apply multi-criteria filtering. We perform frequency-domain analysis to prioritize frames with high sharpness and minimal motion blur, and employ a 3D object grounding model to detect manipulable objects in each candidate frame. Frames exhibiting both high visual clarity and rich object content are retained for subsequent processing.

\noindent\textbf{3D Scene Reconstruction and Operation Generation.}
For each selected image $\mathcal{I}$, we leverage DetAny3D~\cite{zhang2025detect}, an open-vocabulary 3D grounding model, to reconstruct the source scene $\mathcal{S}_{\text{src}} = g(\mathcal{I})$. This extracts object-level 3D bounding boxes, poses, and semantic labels in the camera coordinate system, with camera intrinsics obtained from dataset metadata. With $\mathcal{S}_{\text{src}}$ established, we generate candidate spatial operations (move, rotate, remove) through a rule-based procedure similar to GSI-Syn: randomly selecting a target object and proposing a plausible transformation $\Phi_{\text{3D}}$ to compute $\mathcal{S}_{\text{dst}}$. However, due to positional uncertainty in 3D grounding and the absence of physics simulation, additional quality control is essential.

\noindent\textbf{Visualization-based Verification and MLLM Gating.}
To filter invalid operations, we employ a visualization-driven validation approach. For each candidate operation, we project both the original bounding box $\mathcal{O}_i$ and transformed bounding box $\mathcal{O}_i'$ onto the image plane, generating side-by-side before-and-after visualizations. An MLLM then serves three critical functions: (1) identifying and discarding physically implausible operations (e.g., collisions, floating objects, out-of-frame placements, severe occlusions), (2) correcting annotation errors such as label-object mismatches, and (3) generating diverse natural language instructions by rewriting template-based captions based on visual context and operation metadata.

\noindent\textbf{Human Review and Refinement.}
As a final quality assurance step, we conduct comprehensive manual review of the entire GSI-Real dataset to identify and correct residual annotation inaccuracies or ambiguous instructions, ensuring high annotation quality and a genuinely challenging testbed for real-world spatial reasoning.

\subsection{Evaluation Protocol}
\label{subsec:evaluation}

To comprehensively assess generative spatial intelligence, we design a multi-faceted evaluation protocol with four core metrics.

\noindent\textbf{Instruction Compliance (IC).}
This binary metric evaluates whether the edited scene satisfies the spatial semantics specified in the instruction (e.g., directional relations, containment). We allow reasonable tolerance rather than strict numerical precision: an operation succeeds if the final object pose falls within a plausible range of the ideal target.

\noindent\textbf{Spatial Accuracy (SA).}
For edits passing compliance, we measure fine-grained geometric precision by computing normalized translation error, relative pose error for multi-object operations, and geodesic rotation error on SO(3). Errors are aggregated into a single continuous accuracy score per sample.

\noindent\textbf{Edit Locality (EL).}
We assess localized editing by computing LPIPS \cite{lpips} on non-target regions between original and edited images, using the projected 3D bounding box as a mask to exclude the edited object. Lower LPIPS scores indicate better consistency of unaffected regions. To ensure higher score indicates better performance, we take $100(1-\mathrm{LPIPS})$ as EL score. Before scoring IC and SA, we apply a \textbf{dataset-specific} locality gate using masked SSIM~\cite{ssim} and LPIPS~\cite{lpips} (stricter on synthetic data than on GSI-Real); full thresholds are in the appendix.



\noindent\textbf{Appearance Consistency (AC).}
We leverage an MLLM (Qwen3-VL-235B) to verify appearance quality. For transformation operations (move, rotate, scale), it checks whether the edited object retains its original visual attributes (category, texture, color). For removal operations, it assesses background inpainting quality, identifying residual artifacts or visual discontinuities.


Detailed definitions and thresholds are in the appendix.

\section{Fine-tuning Unified MLLMs for GSI}
\label{sec:fine-tuning}

Beyond evaluation, GSI-Syn's automated synthesis pipeline enables us to construct large-scale editing training data for fine-tuning unified multimodal large language models. 
This allows us to explore two key questions: (1) whether generative training can directly enhance spatial understanding capabilities, and (2) whether unified models can effectively bridge the sim-to-real gap through joint perception-generation learning. 
We choose BAGEL~\cite{deng2025emerging} as our base model, which natively supports image editing and employs self-attention for deep interaction between perception and generation modules, potentially enabling mutual reinforcement between understanding and generation. 
We construct a training set from GSI-Syn comprising diverse spatial operations (move, rotate, resize, remove, scaling, view change)
Further training details are provided in the appendix.


%% file: sec/4_experiments.tex
\section{Experiments}
\label{sec:experiments}

\begin{table*}[t]
\centering
\scriptsize
\setlength{\tabcolsep}{4pt}
\renewcommand{\arraystretch}{1.2}
\caption{\textbf{Performance comparison on the proposed GSI-Bench across three datasets} and four spatial reasoning dimensions: Instruction Compliance (IC), Spatial Accuracy (SA), Appearance Consistency (AC), and Edit Locality (EL). Higher is better.}
\begin{tabular}{lc|cc|cccccccc|c}
\toprule
 & & \multicolumn{2}{c|}{\textbf{Closed-Source Models}} & \multicolumn{8}{c|}{\textbf{Open-Source Models}} &  \\ \cline{3-5} \cline{6-13}
 & \multirow{-2}{*}{\textbf{Evaluation Dimension}}  
 & \textbf{Nano Banana} & \textbf{GPT img} 
 & \textbf{Anyedit} & \textbf{Uniworld} & \textbf{Ultra} & \textbf{Qwen} & \textbf{Omnigen2} & \textbf{Emu3.5} & \textbf{BAGEL} & \textbf{BAGEL+GSI-Syn} & \textbf{$\Delta\uparrow$}\\ 
\hline

\multirow{5}{*}{\rotatebox[origin=c]{90}{\textit{\textbf{GSI-real}}}}
 & IC & 38.78 & 41.72 & 10.20 & 28.80 & 10.66 & \underline{51.02} & 33.56 & \textbf{51.70} & 31.97 & 40.14 & +8.16 \\
 & SA & 21.60 & 28.04 & 8.37 & 18.36 & 5.70 & \textbf{31.22} & 19.62 & \underline{29.51} & 22.07 & 27.76 & +5.68 \\
 & AC & 38.78 & 41.52 & 9.68 & 28.75 & 9.48 & \underline{50.95} & 33.20 & \textbf{51.70} & 31.88 & 40.14 & +8.25 \\
 & EL & 34.92 & 27.52 & 8.75 & 18.51 & 8.97 & \underline{40.55} & 29.82 & \textbf{41.17} & 27.89 & 37.11 & +9.22 \\
 & \textbf{Avg} & 33.52 & 34.70 & 9.25 & 23.61 & 8.70 & \underline{43.44} & 29.05 & \textbf{43.52} & 28.46 & 36.28 & +7.83 \\
\hline

\multirow{5}{*}{\rotatebox[origin=c]{90}{\textit{\textbf{GSI-syn-table}}}}
 & IC & 36.62 & \underline{39.33} & 10.33 & 15.83 & 2.17 & 27.33 & 0.00 & 39.17 & 27.17 & \textbf{50.67} & +23.50 \\
 & SA & \underline{38.96} & 26.16 & 22.84 & 30.33 & 3.09 & 25.52 & 0.00 & 24.09 & 26.52 & \textbf{44.10} & +17.58 \\
 & AC & 36.62 & 38.40 & 10.33 & 15.58 & 1.33 & 27.27 & 0.00 & \underline{38.82} & 26.52 & \textbf{50.67} & +24.15 \\
 & EL & \underline{35.91} & 31.98 & 9.52 & 14.43 & 1.93 & 25.51 & 0.00 & 34.91 & 26.17 & \textbf{49.52} & +23.36 \\
 & \textbf{Avg} & \underline{37.03} & 33.97 & 13.26 & 19.04 & 2.13 & 26.41 & 0.00 & 34.25 & 26.59 & \textbf{48.74} & +22.15 \\
\hline


\multirow{5}{*}{\rotatebox[origin=c]{90}{\textit{\textbf{GSI-syn-room}}}}
 & IC & 20.65 & 8.05 & 7.00 & 12.69 & 2.20 & 20.40 & 18.71 & \underline{20.70} & 16.11 & \textbf{24.01} & +7.90 \\
 & SA & 16.85 & 8.05 & 6.46 & 11.55 & 2.21 & \underline{17.73} & 15.03 & 16.56 & 14.53 & \textbf{19.41} & +4.88 \\
 & AC & 28.01 & 16.69 & 11.85 & 20.40 & 3.46 & \underline{28.67} & 25.94 & 26.98 & 24.00 & \textbf{31.64} & +7.64 \\
 & EL & \underline{19.65} & 7.34 & 5.50 & 11.03 & 1.86 & 18.48 & 17.13 & 17.56 & 14.82 & \textbf{22.61} & +7.79 \\
 & \textbf{Avg} & 21.29 & 10.03 & 7.70 & 13.92 & 2.43 & \underline{21.32} & 19.20 & 20.45 & 17.37 & \textbf{24.42} & +7.05 \\
\bottomrule
\end{tabular}
\end{table*}
\subsection{Experimental Setup}

\noindent\textbf{Benchmarks and Dataset Statistics.}
Our evaluation suite consists of two complementary benchmarks. 
\textbf{GSI-Real} contains 441 samples from 211 diverse indoor scenes in ScanNet++~\cite{Scannet++}, spanning three operation types.
\textbf{GSI-Syn} comprises two subsets: \emph{GSI-Syn-Room} (593 samples, six operations) built on AI2-THOR~\cite{Ai2-thor}, and \emph{GSI-Syn-Tabletop} (600 samples, three operations) using MesaTask~\cite{mesatask}.
Dataset statistics are in Figure~\ref{fig:main}.
To evaluate cross-view generalization, we construct \emph{GSI-Syn-Bathroom} with 200 samples featuring randomized viewpoints.
For fine-tuning, \textbf{GSI-Syn-Train} contains 1,500 training samples per operation type per environment, totaling 10,500 samples with strict scene separation from test sets.


\noindent\textbf{Baseline Models.} We evaluate nine state-of-the-art models: seven open-source models (BAGEL~\cite{deng2025emerging}, Anyedit~\cite{jiang2025anyedit}, Uniworld~\cite{lin2025uniworld}, Ultra~\cite{zhao2024ultraedit}, Qwen-Image-Edit~\cite{wu2025qwen}, Omnigen2~\cite{wu2025omnigen2}, Emu3.5~\cite{cui2025emu3}) and two proprietary models (NanoBanana~\cite{comanici2025gemini}, GPT-image~\cite{openai2024gpt4o}). These models span diverse architectures including unified MLLMs and instruction-based editors, evaluated using publicly available checkpoints or API endpoints with default settings.

\subsection{Benchmarking Generative Spatial Intelligence}

Across GSI-Bench, we observe clear performance disparities reflecting different levels of spatial reasoning capability.

\noindent\textbf{Closed-Source Models.}
On GSI-Syn-Table, Nano Banana and GPT-img reach 37.03 and 33.97 average respectively, with strength in IC and AC. On GSI-Real, however, their averages (33.52 and 34.70) are only comparable to open-source systems like Qwen (43.44) and Emu3.5 (43.52), indicating that closed-source models, despite strong general visual generation, struggle with fine-grained spatial manipulations requiring explicit geometric understanding.

\noindent\textbf{Open-Source Baselines.}
Emu3.5 is the strongest open-source performer, achieving the best results on GSI-Real (43.52 average) and high scores across all dimensions. In contrast, general-purpose models like Uniworld, Ultra, and Omnigen2 show substantially lower scores, with extremely low AC or IC values revealing difficulty following structured spatial instructions. These results suggest most open-source models lack 3D-aware inductive biases for precise spatial reasoning, whereas Emu3.5 benefits from stronger spatial priors through its video-centric training.

Qualitative results (Fig~\ref{fig:vis} and more in appendix) reveal consistent trends: most models perform better on removal than other operations, indicating deletion is easier than precise geometric manipulation. Emu3.5 produces the cleanest removals with strongest spatial consistency. However, Ultra and AnyEdit often fail to preserve object identity; AnyEdit, BAGEL, and Omnigen2 introduce artifacts; AnyEdit frequently leaves targets unchanged; BAGEL sometimes misinterprets translation as camera motion. While BAGEL, Emu3.5, and Qwen reliably follow referential cues, they occasionally remove additional content, indicating fine-grained localization remains challenging.

\begin{table}[t]
\centering
\small
\renewcommand{\arraystretch}{1.2}
\setlength{\tabcolsep}{3.5pt}
\caption{
\textbf{Evaluation on OmniSpatial benchmark.} 
We report accuracy (\%) across four core reasoning dimensions. 
Fine-tuning on GSI-Syn improves spatial understanding, particularly in Spatial Interaction and Perspective Taking.
Best results among open-source 7B models are \textbf{bolded}. 
\textsuperscript{\dag}Proprietary models.
}
\label{tab:omnispatial_summary}
\resizebox{0.9\columnwidth}{!}{
\begin{tabular}{lccccccc}
\toprule
\textbf{Model} & \textbf{Size} & \textbf{Overall} & \textbf{Dynamic} & \textbf{Spatial} & \textbf{Logic} & \textbf{Persp.} \\
\midrule
GPT-4-turbo\textsuperscript{\dag}~\cite{achiam2023gpt4} & -- & 34.06 & 38.39 & 36.49 & 24.80 & 33.69 \\
Gemini-2.5\textsuperscript{\dag}~\cite{gemini2023anil} & -- & 52.12 & 63.59 & 67.46 & 35.67 & 43.10 \\
\midrule
LLaVA-1.5~\cite{liu2024llava15} & 7B & 34.97 & 35.38 & 35.13 & 25.99 & 38.82 \\
Qwen-VL-2.5~\cite{wang2024qwen2vl} & 7B & 39.25 & 46.30 & 30.06 & \textbf{35.65} & 39.68 \\
BAGEL~\cite{deng2025emerging} & 7B & 41.55 & 47.38 & 45.67 & 32.14 & 39.22 \\
\midrule
\textbf{BAGEL + GSI-Syn} & 7B & \textbf{42.07} & \textbf{48.33} & \textbf{47.67} & 28.97 & \textbf{40.29} \\
\bottomrule
\end{tabular}
}
\vspace{-2mm}
\end{table}

\begin{table}[t]
\centering
\small
\renewcommand{\arraystretch}{1.2}
\setlength{\tabcolsep}{4.5pt}
\caption{
\textbf{Evaluation on SAT-Real benchmark~\cite{ray2024sat}.}
Accuracy (\%) across five spatial reasoning dimensions. 
Fine-tuning with GSI-Syn notably improves goal-directed and egocentric understanding.
Best results among open-source 7B models are \textbf{bolded}.
}
\label{tab:satreal_summary}
\resizebox{0.9\columnwidth}{!}{
\begin{tabular}{lcccccc}
\toprule
\textbf{Model} & \textbf{Overall} & \textbf{Pers} & \textbf{GoalAim} & \textbf{EgoAct} & \textbf{ObjM} & \textbf{EgoM} \\
\midrule
Qwen-VL-2.5~\cite{wang2024qwen2vl} & 56.33 & 43.94 & 67.65 & 56.76 & 56.52 & 56.52 \\
BAGEL~\cite{deng2025emerging} & 65.33 & 46.97 & 75.00 & \textbf{75.68} & 65.22 & 60.87 \\
\textbf{BAGEL + GSI-Syn} & \textbf{69.33} & \textbf{48.48} & \textbf{85.29} & {72.97} & \textbf{65.22} & \textbf{73.91} \\
\bottomrule
\end{tabular}
}
\vspace{-2mm}
\end{table}

\begin{figure*}[htbp]
    \centering
    \includegraphics[width=0.9\linewidth]{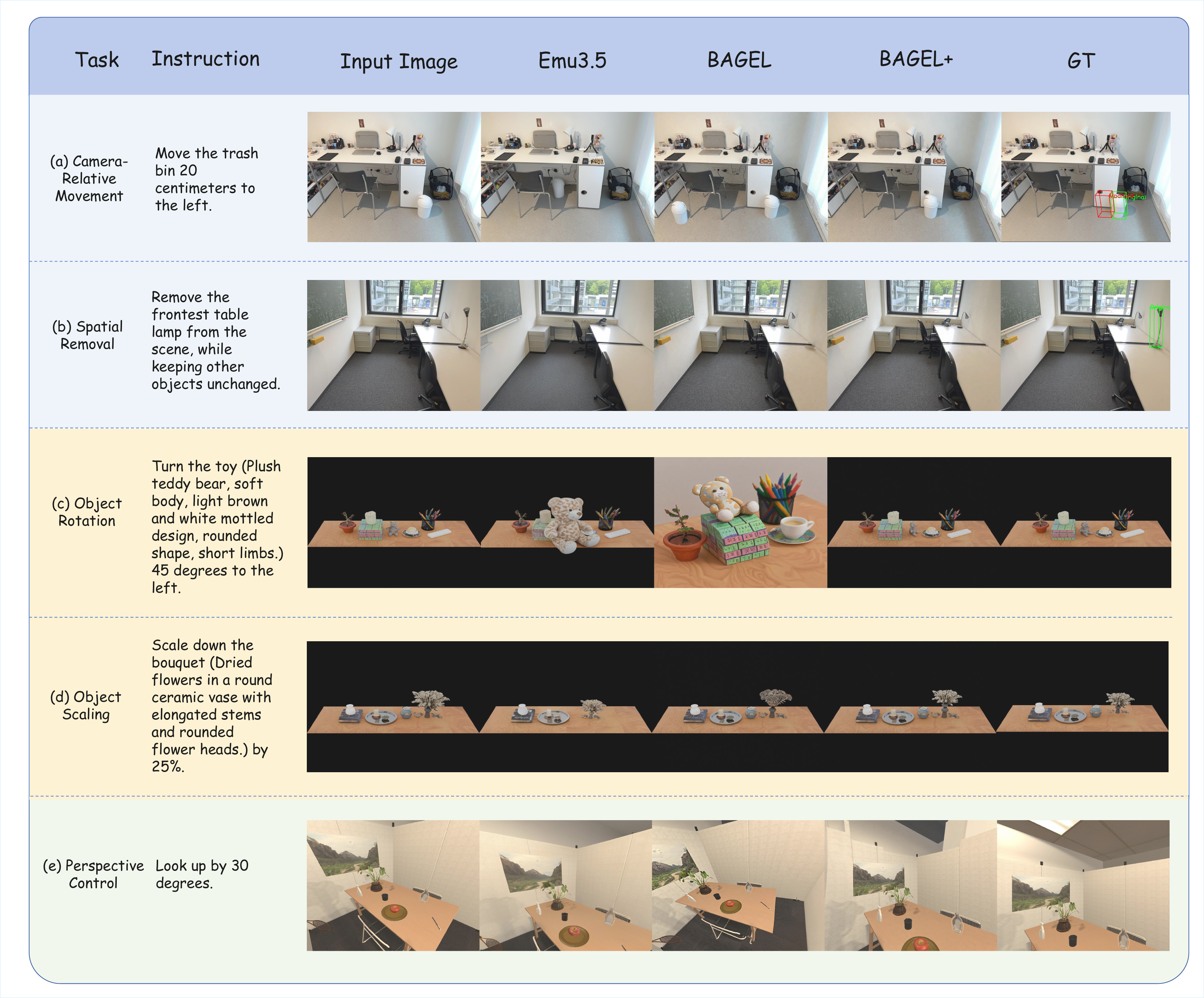}
    \caption{Qualitative comparison of spatial editing results across five instruction types.
Rows 1–2 use GSI-Real samples, Rows 3–4 use GSI-Table, and the last row uses GSI-Room.
Columns show the input image, outputs from Emu3.5, BAGEL, BAGEL+(fine-tuned with GSI-Syn), and the ground-truth target.
BAGEL+ demonstrates stronger spatial fidelity and better preservation of unaffected content.
Further examples and corresponding metrics are provided in the appendix.}
\label{fig:vis}
\vspace{-2mm}
\end{figure*}

\subsection{Impact of Fine-tuning on GSI-Syn}

\noindent\textbf{Effective Sim-to-Real Transfer.}
Fine-tuning on GSI-Syn yields consistent improvements across both domains. 
On GSI-Real, the model achieves a 7.83-point average gain over BAGEL (28.46$\rightarrow$36.28). The largest gains are in Edit Locality (+9.22), Appearance Consistency (+8.25), and Instruction Compliance (+8.16), indicating better preservation of object identity and more precise, spatially constrained edits despite training exclusively on synthetic images; Spatial Accuracy also improves (+5.68).
On synthetic benchmarks, improvements are even larger: +22.15 on GSI-Syn-Table and +7.05 on GSI-Syn-Room. The model benefits particularly from the structured geometric variations in GSI-Syn-Table targeting localized edits. Gains on GSI-Syn-Room are more modest due to increased scene complexity and spatial ambiguities, highlighting remaining limitations in global spatial reasoning.
These results demonstrate that geometrically grounded synthetic supervision significantly enhances spatial editing capabilities and transfers robustly to real images without requiring real-world annotations.

\noindent\textbf{Enhanced Spatial Understanding through Generative Training.} 
As shown in Table~\ref{tab:omnispatial_summary}, fine-tuning BAGEL solely on spatially-related generative editing data (GSI-Syn)---without any understanding or reasoning data---improves performance on the OmniSpatial benchmark. 
BAGEL shows consistent gains in the most relevant dimensions: \textit{Dynamic Reasoning} (+0.95\%), \textit{Spatial Interaction} (+2.00\%), and \textit{Perspective Taking} (+1.07\%). 
We observe a moderate decrease in \textit{Complex Logic}, attributable to the absence of explicit reasoning supervision in the fine-tuning corpus. 
Nevertheless, the overall improvement provides strong evidence that generative spatial training alone substantially enhances spatial understanding, highlighting a promising direction for unified MLLMs that jointly leverage generative and reasoning-based objectives.
Results on SAT-Real~\cite{ray2024sat} (Table~\ref{tab:satreal_summary}) further validate this finding: fine-tuning on GSI-Syn yields notable improvements in \textit{Allocentric Perspective}, \textit{Goal Aiming}, and \textit{Egocentric Movement}, achieving an overall gain of +4.00\%.

%% file: sec/5_conclusion.tex
\section{Conclusion}
\label{sec:Conclusion}
This paper studies Generative Spatial Intelligence. We introduce GSI-Bench, a benchmark spanning seven spatial operation categories, with a real-world set, a large-scale synthetic set, and automated pipelines based on 3D grounding priors.
Experiments show that current state-of-the-art models still struggle with spatially accurate generation. Fine-tuning on GSI-Syn improves spatial compliance and transfers to real-world and spatial understanding tasks, suggesting that generative training enhances spatial reasoning.